\documentclass[11pt]{article}

\usepackage{ACL2023}

\usepackage{times}
\usepackage{latexsym}
\usepackage{graphicx} 
\usepackage{multirow}
\usepackage{amsmath}
\usepackage{arydshln}
\usepackage{xcolor}
\usepackage[T1]{fontenc}
\usepackage[utf8]{inputenc}

\usepackage{microtype}

\usepackage{inconsolata}
\usepackage[caption=false]{subfig}
\usepackage{pgfplots} 
\usepackage{adjustbox}
\title{ParaLS: Lexical Substitution via Pretrained Paraphraser}

 \author{Jipeng Qiang \and Kang Liu \and Yun Li  \and  Yunhao Yuan \and Yi Zhu   \\
    School of Information Engineering, Yangzhou University, Yangzhou, China
    \\
    \texttt{\{jpqiang, liyun, yhyuan, zhuyi\}@yzu.edu.cn}
    }
\begin{document}
\maketitle
\begin{abstract}
Lexical substitution (LS) aims at finding appropriate substitutes for a target word in a sentence. Recently,  LS methods based on pretrained language models have made remarkable progress, generating potential substitutes for a target word through analysis of its contextual surroundings. However, these methods tend to overlook the preservation of the sentence's meaning when generating the substitutes. This study explores how to generate the substitute candidates from a paraphraser, as the generated paraphrases from a paraphraser contain variations in word choice and preserve the sentence's meaning. Since we cannot directly generate the substitutes via commonly used decoding strategies, we propose two simple decoding strategies that focus on the variations of the target word during decoding. Experimental results show that our methods outperform state-of-the-art LS methods based on pre-trained language models on three benchmarks.
\end{abstract}

\section{Introduction}
Lexical substitution (LS) in context \cite{hintz2016language,zhou2019bert,arefyev2020comparative} is an extremely powerful technology that can be used as a backbone of various NLP applications such as writing assistance \cite{lee-etal-2021-swords}, word sense disambiguation \cite{mccarthy2002lexical}, and lexical simplification \cite{paetzold2016unsupervised,qiang2021lsbert,qiang2021chinese}. Compared with traditional 
LS methods based on linguistic databases (e.g., WordNet) \cite{hassan2007unt,yuret2007ku} or word embedding models \cite{melamud2015modeling,melamud2015simple}, LS methods based on pretrained language models have made remarkable progress in generating substitutes by considering the context \cite{zhou2019bert,qiang2021lsbert,michalopoulos2022lexsubcon,seneviratne-etal-2022-cilex}. These methods feed the sentence into BERT \cite{devlin2018bert} or XLNet \cite{yang2019xlnet} to obtain the top probability words corresponding to the target word as the substitute candidates. However, they have the following two limitations. 

(1) The predictability of words is greatly influenced by the surrounding context, with little regard for preserving the sentence's meaning. As illustrated in Table 1, the utilization of pretrained models often leads to the generation of ill-suited words, such as "wet", "flat" and "cold", due to their contextual relevance and similarity to the target word.
 
(2) The utilization of subword techniques in pretrained models precludes the selection of multi-token words as substitutes, as they only generate the most probable single tokens. For instance, the words "desiccated" and "facilitated" would not be offered as a substitution for the target word "dry" as seen in Table 1.

\begin{table}
\centering
\begin{tabular}{l|l}
\hline

\multirow{2}{*}{\textbf{Sent1}} &  surprisingly in such a \textbf{dry} continent  \\
 & as Australia , salt becomes a  $\cdot\cdot\cdot$ \\ \cdashline{1-2} 

\textbf{Labels}  & arid, waterless \\
\cdashline{1-2} 
 \textbf{BERT} & wet,\textcolor{red}{arid},moist,humid,damp\\
 \cdashline{1-2} 
 \textbf{XLNet} & wet, flat, moist, \textcolor{red}{desert}, cold \\
 \cdashline{1-2} 
 \multirow{2}{*}{\textbf{Ours}} & \textcolor{red}{desiccated},\textcolor{red}{drought},\textcolor{red}{arid},dead, \\
 & \textcolor{red}{parched} \\
\hline
\multirow{2}{*}{\textbf{Sent2}} &  remember that the delegates ' life    \\
& is not always \textbf{easy}. \\ 
\cdashline{1-2} 
\multirow{2}{*}{\textbf{Labels}}  & simple, trouble free, undemanding, \\
& uncomplicated, straightforward \\
\cdashline{1-2} 
\multirow{2}{*}{ \textbf{BERT} } &\textcolor{red}{simple}, hard,complicated,\\
 & difficult, exciting  \\
 \cdashline{1-2} 
 \textbf{XLNet} & cheap, \textcolor{red}{simple}, quick, hard, fast \\
 \cdashline{1-2} 
 \multirow{2}{*}{\textbf{Ours}} & \textcolor{red}{simple}, light, \textcolor{red}{good}, \textcolor{red}{ease}, \\
 & \textcolor{red}{facilitated} \\
\hline
\end{tabular}
\caption{The top five substitutes of two instances in LS07 dataset using BERT \cite{zhou2019bert}, XLNet \cite{seneviratne-etal-2022-cilex} and our method. The target word of each sentence is bolded, and the suitable substitutes are marked in red. }
\label{sample1}
\end{table}

To address the limitations mentioned above, we study how to generate substitutes via paraphrase modeling. Recent neural paraphrasers based on encoder-decoder framework \cite{wieting2017paranmt,hu-etal-2019-large} produce fluent, meaning-preserving English paraphrases but contain variations in word choice. Therefore, our idea is whether we can decode the substitute candidates from the hidden representation of the target word. In this way, the substitutes are not only semantically consistent with the target word and fit in the context, but also can preserve the sentence’s meaning. The meaning-preserving properties of a paraphraser can aid in addressing the first limitation, while autoregressive paraphrasers can address the second limitation. To the best of our knowledge, paraphraser for LS task has not yet been explored, as current decoding methods focus on lexical variations within the entire sentence rather than the target word, resulting in a scarcity of appropriate substitutes for the target word.

To specifically focus on lexical variations of the target word during the decoding process, we propose two new decoding strategies. (1) Our first strategy, referred to as ParaLS, proposes fixing prefixes of the target word. This approach initiates the decoding process by mandating that the decoder begins with the target word's prefixes in the sentence, to subsequently generate the probability distribution of the target word's position. The words with the highest probabilities are then fixed and used when decoding the remaining words, with selected words of the target word's position in the paraphrases being selected as substitute candidates.
(2) The second strategy, referred to as ParaLS$\star$, is proposed to address the oversight of suffixes in the first strategy. Inspired by NEUROLOGIC A$\star$esque \cite{lu-etal-2022-neurologic}, which incorporates heuristic estimates of future cost, we adapt it to estimate of the words in suffixes.

To the best of our knowledge, ParaLS is the first LS method that can produce substitute candidates by considering preserving the sentence's meaning. On three benchmarks, ParaLS and ParaLS$\star$ achieve state-of-the-art performance across various evaluation metrics. Moreover, ParaLS$\star$ without the step of substitute ranking outperforms all existing methods with the step of substitute ranking.
 
Additionally, we propose a novel strategy for the step of substitute ranking by text generation evaluation metrics BARTScore \cite{NEURIPS2021_e4d2b6e6} and BLEURT \cite{sellam2020bleurt}. Our method embeds each substitute into the original sentence to create an updated version. By using BARTScore and BLEURT to compute the relationship between the original and updated sentences, they can quantify the extent to which the meaning of the original sentence has been preserved by each substitute. Experimental results show that substitute ranking using only BARTScore outperforms the previous state-of-the-art ranking methods when the same substitution candidate lists are provided for two popular LS benchmarks. The code and the experimental results are source-opened in Github \footnote{https://github.com/qiang2100/ParaLS}.

\section{Related Work}

\textbf{Lexical Substitution.} LS methods generally consist of two steps: substitute generation and substitute ranking. Previous LS methods utilize linguistic databases (e.g., WordNet) \cite{hassan2007unt,yuret2007ku} or
word embedding models \cite{melamud2015simple, melamud2015modeling, qiang2019unsupervised} to extract synonyms or highly similar words for a target word, and then sort them based on their appropriateness in context. These methods overlook the context of the target word while generating substitute candidates, thereby inevitably generating a plethora of irrelevant candidates that may impede the subsequent ranking phase.

Recent LS methods based on pretrained language models have attracted much attention \cite{zhou2019bert,lacerra2021alasca,michalopoulos2022lexsubcon}, in which the pretrained BERT is the most widely used one. Zhou et al. \cite{zhou2019bert} apply dropout to the embeddings of target words, Michalopoulos et al. \cite{michalopoulos2022lexsubcon} propose a new mix-up embedding strategy by incorporating the knowledge of WordNet into the prediction process of BERT, and Lin et al. \cite{lin-etal-2022-improving} proposed an auxiliary gloss regularizer module to BERT pre-training. Lacerra et al. \cite{lacerra2021genesis,lacerra2021alasca} tried to train pretrained language models by merging the development set of two LS datasets (CoInCo and TWSI). The current work \cite{arefyev2020comparative,seneviratne-etal-2022-cilex} sought to evaluate all existing pretrained language models, and found that combining the prediction of pretrained language models XLNet and Word2Vec achieved the best results. 

Overall, pretrained language modeling-based LS methods consider contextual information of target words when generating substitute candidates, but do not concern with the impact of applying substitutes on sentence meaning. In contrast to the aforementioned methods, we try to utilize the knowledge of a pretrained paraphraser to generate substitute candidates.

\textbf{Lexical Substitution using Paraphrases.} A few studies \cite{pavlick2016simple,Kriz2018Simplification} find substitute candidates for complex words from a large-scale paraphrase rule database, e.g., PPDB \cite{Ganitkevitch2013} or its variations \cite{pavlick2015ppdb,pavlick2016simple}. A paraphrase rule database consists of large-scale lexical paraphrase rules (e.g., "berries$\rightarrow$strawberries") that are extracted from large-scale paraphrase sentence pairs, such as ParaNMT \cite{wieting2017paranmt} or ParaBank \cite{hu-etal-2019-large}. These methods do not take into account the context as linguistic resource-based LS methods do. In this paper, we generate substitute candidates of target words using the pretrained paraphrase model instead of paraphrase rule databases or paraphrase databases.

\textbf{Decoding Strategies.} Paraphrase generation can be regarded as a monolingual machine translation task that transforms expressions of an input sentence while retaining its meaning \cite{wieting2017paranmt}. Neural paraphrasers primarily rely on the encoder-decoder framework, achieving inspiring performance gains over the traditional approaches \cite{lu-etal-2022-neurologic}. Beam search decoding is the most common method for inference, which decodes the top-$K$ sequences in a greedy left-to-right fashion. When $K$ is set to 1, beam search decoding is changed into greedy search decoding. In recent years, beam search decoding has had multiple variants to deal with various task-specific and diversity/fluency trade-off of outputs, such as noise beam decoding \cite{cho2016noisy}, iterative beam decoding \cite{Kulikov2019}, clustered beam decoding \cite{tam2020cluster} and diverse beam decoding \cite{vijayakumar2018diverse}. To enable
constrained generation, NEUROLOGIC A$\star$esque \cite{lu-etal-2022-neurologic} explicitly decodes future text to estimate the viability of different paths for satisfying constraints. In contrast to the above decoding strategies, our decoding strategies focus solely on enhancing the diversity of the target word's variation.

\section{Method}

Given a given sentence $\mathbf{x}=\{x_1,x_2,...,x_t, ..., x_n\}$ and the target word $x_t$, we need a pretrained paraphraser based on an autoregressive model, instead of a pretrained language modeling like existing LS methods \cite{zhou2019bert,michalopoulos2022lexsubcon,seneviratne-etal-2022-cilex}, e.g., BERT or XLNet. LS method consists of two steps: substitute generation and substitute ranking. After feeding sentence $\mathbf{x}$ into the paraphraser, we aim to extract substitute candidates for the target word $x_t$ by two novel decoding strategies (Section 2.2). Then, we rank the candidates to choose the most appropriate substitution without modifying the meaning of $\mathbf{x}$ (Section 2.3).

\subsection{Motivation}

\begin{figure}
  \centering
  {
     \includegraphics[width=78mm]{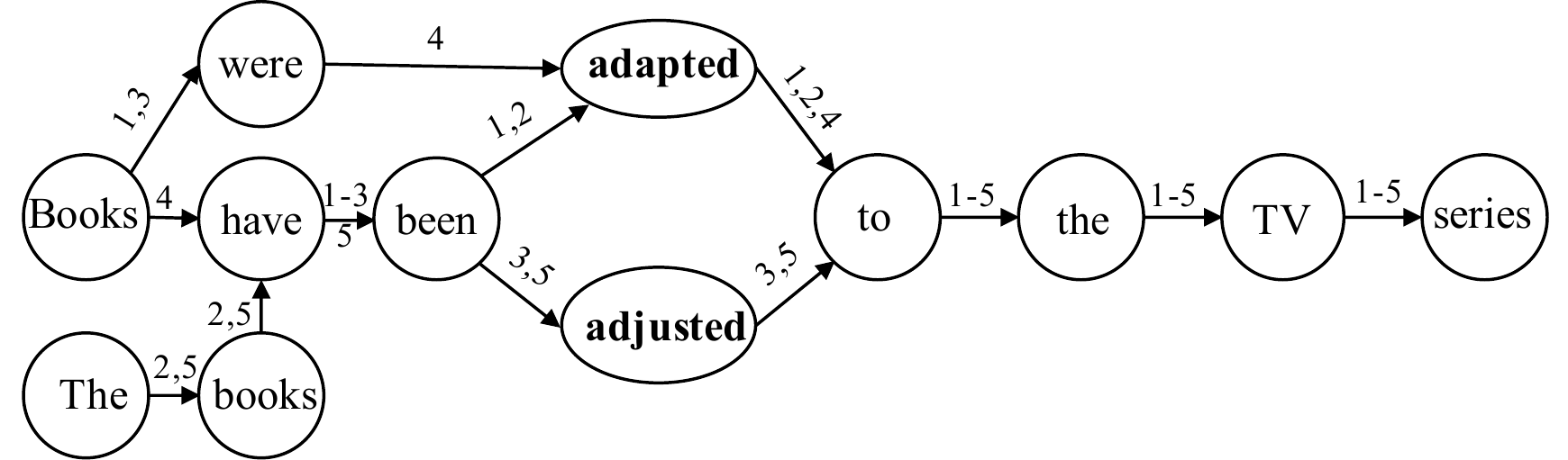} \\
     (a) Five paraphrases using beam decoding.
  }
  {
     \includegraphics[width=78mm]{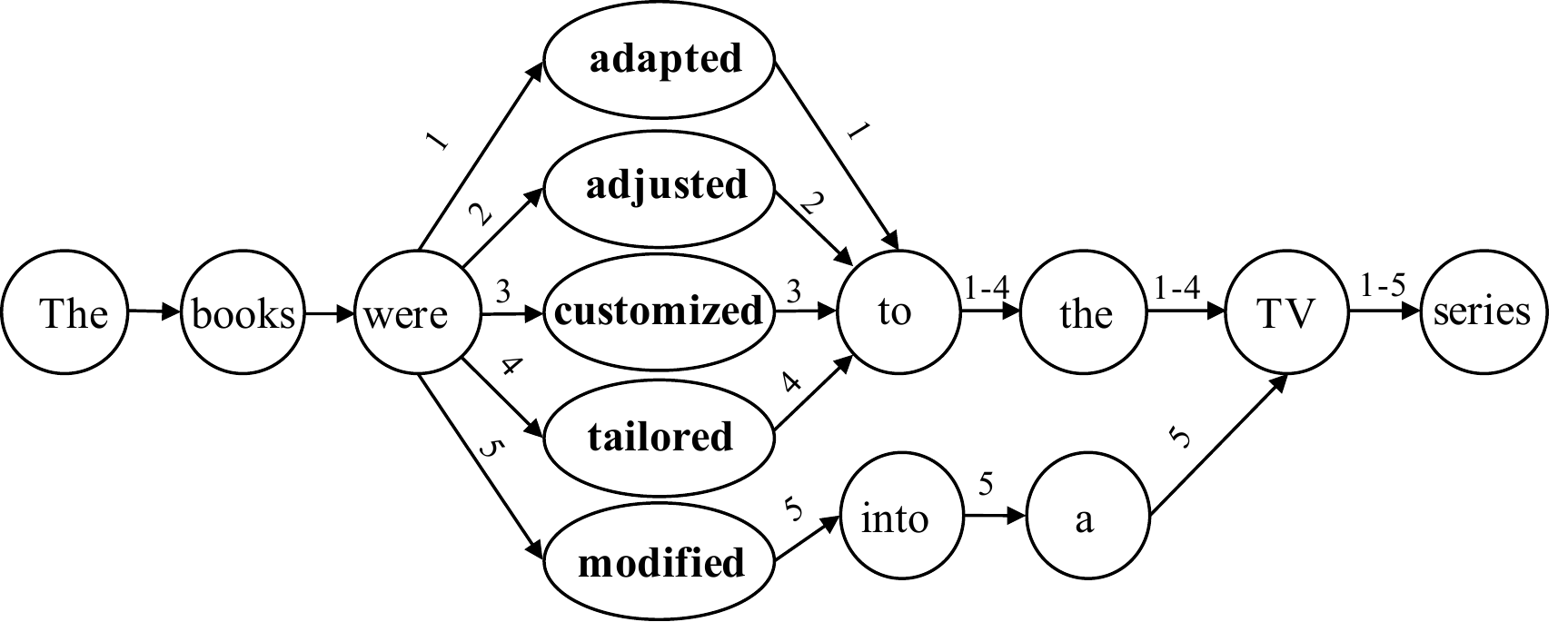} \\
     (b) Five paraphrases using our decoding strategy via fixing prefixes.
  }
 \caption{The paraphrases of the sentence "The books were adapted into a television series" using two different decoding methods. "adapted" is the hypothetical target word.  Figure (a) shows the normal paraphrases of beam decoding with beam size 5, and Figure (b) shows the first 5 paraphrases of our decoding method by forcing the decoder to begin with prefixes "The books were" of the target word. We have easily access to substitute candidates of the target word "adapted" from the paraphrases using our decoding method. }
 \label{Fig1}
\end{figure}

Recent neural paraphrasers primarily rely on the encoder-decoder learning framework on a large-scale paraphrase dataset, achieving inspiring performance gains over traditional methods \cite{meng-etal-2021-conrpg,kadotani-etal-2021-edit}. Many languages including English, French, German, Chinese, and Spanish own large-scale paraphrase datasets. For example,  in English, ParaBank2 \cite{hu-etal-2019-large} consists of 19,370,798 sentence pairs.

Given an input sentence $\mathbf{x}$ and its corresponding paraphrase \textbf{y}, we consider standard left-to-right, autoregressive models, $p_\theta(\textbf{y}|\textbf{x})=\prod_{t=1}^{|\textbf{y}|}p_\theta(y_t|\textbf{y}_{<t},\textbf{x})$, and omit \textbf{x} to reduce clutter. Decoding consists of solving,

\begin{equation}
 \textbf{y}_* = \mathop{\arg\max}_{Y\subseteq \mathcal{Y}} F(\textbf{y})
\end{equation}
where $\mathcal{Y}$ is the set of all sequences, and the objective $F(\textbf{y})$ is $\log p_{\theta}(\textbf{y})$.

If we want to generate multiple paraphrases of sentence $\mathbf{x}$, the beam search decoding is widely used by the auto-regressive method, which maintains a beam of $K$ possible generations, updating them incrementally by ranking their extensions via the model likelihood. Since beam search decoding aims to find the most-probable hypothesis for the whole sentence during decoding, it is difficult to extract multiple substitute candidates for the target word from the generated paraphrases,  as shown in Figure 1(a). 

Since beam decoding concerns lexical variations of the whole sentence instead of the target word, there are no sufficient appropriate substitutions that can be discovered for the target word if we directly extract the substitute candidates from the paraphrases using the beam decoding. We will propose two novel decoding strategies, ParaLS and ParaLS$\star$, for the paraphraser that are specifically engineered to harness lexical variations of the target word.

\subsection{Substitute Generation}

Substitute generation aims to generate substitute candidates for the target word based on its context. We will generate the candidates during the process of decoding. 

The process of decoding method can be treated as a discrete search, in which $states$ are partial prefixes, $\mathbf{y}_{<t}$, $actions$ are tokens in vocabulary $\mathcal{V}$ (i.e., $y_t\in \mathcal{V}$), and $transitions$ add a token to prefixes, $\mathbf{y}_{<t} \circ y_t$. Each step of decoding consists of (1) expanding a set of candidate next-states, (2) scoring each candidate, and (3) selecting the best $K$ candidates.

(1) \textbf{ParaLS: Decoding by Fixing Prefixes.} Given a sentence $\mathbf{x}$ and a target word $x_t$, we force the decoder to begin with prefixes $\mathbf{x}_{<t}$ of the target word, and decode succeeding token $y_t$ to estimate the probability distribution of the vocabulary $p(y_t|\mathbf{x}_{<t})$. We select the top $K$ tokens $Y_t$ with the highest probability in the distribution as the results of decoding.

\begin{equation*}
Y_t^{'}=\{\mathbf{y}_{<t}\circ y_t | \mathbf{y}_{<t}=\mathbf{x}_{<t}, y_t \in \mathcal{V}\}
\end{equation*}
\begin{equation}
Y_t = \mathop{\arg topK}_{(\mathbf{y}_{<t}\circ y_t)\in Y_t^{'}} \{f(\mathbf{y}_{<t},y_t)\}
\end{equation}
where $f(\cdot)$ is scoring function that approximates the objective $F$. 

The decoding phase by fixing prefixes $\mathbf{x}_{<t}$ is crucial to generate substitute candidates since we forcibly generate $K$ different tokens $Y_t$ with the highest probability. In this case, these generated tokens are not only semantically consistent with the target word and fit in the context, but also preserve the sentence's meaning. Since one word may comprise two or more tokens, we adopt greedy search decoding to select the token that has the maximum probability for each preceding token, until reaching the end symbol "EOS" of one sentence. These $K$ words are considered as substitute candidates, after eliminating the morphological derivations of the target word. As depicted in Figure 1(b), our decoding strategy concentrates on lexical variations of the target word.

(2) \textbf{ParaLS$\star$:Decoding with Lookahead Heuristics.}  ParaLS, by fixing prefixes, takes into account only prefixes $\mathbf{x}_{<t}$ without accounting for suffixes $\mathbf{x}_{>t}$. In this manner, the top $K$ tokens $Y_t$ by Equation (2) may be one word of suffixes. Drawing inspiration from the A$^\ast$ search algorithm \cite{hart1968formal} and NEUROLOGIC A$^\star$esque \cite{lu-etal-2022-neurologic}, ParaLS$\star$ will incorporate an estimate of the words of suffixes into the prediction of $p(y_t|\mathbf{x}_{<t})$, replacing Equation 2 with:
\begin{equation}
Y_t = \mathop{\arg topK}_{(\mathbf{y}_{<t}\circ y_t)\in Y_t^{'}} \{max F(\mathbf{y}_{<t},y_t, \mathbf{y}_{>t}\}
\end{equation}
where $\mathbf{x}_{>t}$ represents  suffixes, $\mathbf{y}_{<t}$ equals $\mathbf{x}_{<t}$, and $\mathbf{y}_{>t}$ equals $\mathbf{x}_{>t}$.

ParaLS$\star$ enhances the ParaLS scoring function by incorporating an estimate of suffixes satisfaction. Our key addition is a lookahead heuristic that adjusts a candidate ($\mathbf{y}_{<t}$,$y_t$)'s score proportional to the probability of satisfying additional suffixes constraints $\mathbf{y}_{>t}$. In reality, we need only estimate two or three words in suffixes without estimating suffixes. 

Intuitively, our lookahead heuristic for decoding brings two benefits. (1) The $y_t$ can be a token that would satisfy a multi-token constraint or a phrase as the lookahead computes the average score  ($y_t$,$\mathbf{y}_{>t}$). (2) When 
$y_t$ is one word in suffixes, the lookahead will help to decrease its score, thereby precluding it from being among the top $K$ tokens.

\subsection{Substitute Ranking}

After obtaining substitute candidates, existing LS methods \cite{zhou2019bert,lacerra2021genesis,seneviratne-etal-2022-cilex} obtain a contextualized representation of each substitute by replacing the target word with the substitute, and rank the substitutes by computing the cosine similarity of the target word vector with respect to that of each substitute. The similarity between the target word and the substitute does not provide sufficient information about whether the substitute will modify the sentence's meaning. After replacing the target word of the original sentence with the substitute to form the updated version, we attempt to evaluate the original sentence $\mathbf{x}$ and the updated sentence to rank the substitutes, as opposed to the target word and the substitute alone.

We formulate evaluating updated sentence as a text generation evaluation task. Assume that the updated sentence is denoted as $\mathbf{x}'$ after replacing the target word $x_t$ in $\mathbf{x}$ into one substitute. To accurately calculate a similarity score between $\mathbf{x}$ and $\mathbf{x}'$, we found BARTScore \cite{NEURIPS2021_e4d2b6e6} and BLEURT \cite{sellam2020bleurt} are specifically designed for text generation tasks, which aligns with the goal of lexical substitution. Therefore, they could be used to measure the quality of the substitutes.

BARTScore is a neural network-based evaluation metric that compares the likelihood of the original sentence and the updated sentence. It can assign higher scores to the sentences that are more likely to be original sentences. BLEURT is also a neural network-based evaluation metric, which is trained to predict how human-like a text is by comparing it with a large dataset of human-written texts. Those two metrics could assign a similarity or dissimilarity score, which allow the ranking of the substitutes based on how much similar to the original sentence they are, which might be better to rank the substitutes rather than other ranking methods \cite{zhou2019bert,lacerra2021genesis,seneviratne-etal-2022-cilex}.

We have also incorporated the prediction scores of the substitute candidates generated by the paraphraser. Ultimately, our method employs a linear combination of the aforementioned three features (Paraphraser, BARTScore, BLEURT) to compute the final score for each substitute candidate.

\section{Experiments}

\subsection{Experiment Setup}

\textbf{LS Benchmarks.} Two widely used datasets, LS07 \cite{mccarthy2007} and CoInCo \cite{kremer2014substitutes}, are chosen for the evaluation of LS methods. We also adopt the latest LS benchmark, Stanford Word Substitution Benchmark (SwordS) \cite{lee-etal-2021-swords}, which extends
and improves CoInCo via crowdsourcing annotators in Amazon Mechanical Turk. Each instance in LS dataset is composed of a target word, its context, and corresponding substitutes. LS07 consists of 300 development examples and 1710 test instances for 201 polysemous words. CoInCo consists of 15K target instances with a given 35\% development and 65\% test. SwordS contains 762 test instances.

\begin{table*}
\centering
\begin{tabular}{l|l|ccccc}\hline
\textbf{Data set} & \textbf{Method} & \textbf{best} & \textbf{best-m} & \textbf{oot} & \textbf{oot-m} & \textbf{P@1} \\\hline
\multirow{7}{*}{LS07} & Word2Vec  & 12.7 & 21.7  & 36.4 & 52.0  & -  \\
& BERT  & 20.3 & 34.2 & 55.4 & 68.4 & 51.1 \\
& BERT+WordNet   & 21.1(16.3) & 35.5(27.6) & 51.3(45.6) & 68.6(62.4) & 51.7 (40.8)\\
& GR-RoBERT & 23.1(19.4) & 39.7(33.2) & 57.6(52.8) & 76.3 (71.5) & 55.0(47.4) \\
& XLNet+Word2Vec   & 23.3(21.3) & 40.9(37.8) & 56.3(55.04) & 74.8(73.9) & 55.9 (50.5)\\
\cdashline{2-7} 
& ParaLS (ours) & 23.5(20.0) & 41.5(34.4) & 59.0(52.4) & 77.9(68.9) & 56.9(48.4) \\
& ParaLS$\star$ (ours)  & \textbf{24.0(22.3)}  & \textbf{42.2(39.0)} & \textbf{60.5(57.3)} & \textbf{79.3(76.1)} & \textbf{58.8(54.3)} \\
\hline
\multirow{7}{*}{CoInCo} &  Word2Vec & 8.1 & 17.4  & 26.7 & 46.2  & -  \\
& BERT & 14.5 & 33.9 & 45.9 & 69.9 & 56.3 \\
& BERT+WordNet   & 14.0 (11.3) & 29.7(23.8) & 38.0(33.6) & 59.2(54.4) & 50.5(41.3) \\
& GR-RoBERT & 15.2(13.1) & 34.4(28.8) & 45.3(40.9) & 71.3(66.6) & 55.9(48.8) \\
& XLNet+Word2Vec & 16.4(15.1) & 35.8(33.0)& 46.9(45.1) & 73.0(71.9)& 57.3(52.6) \\
\cdashline{2-7} 
& ParaLS(ours) & 18.1(13.8) & 40.1(29.5) & 50.7(41.7) & 78.1(65.6) & 62.4(50.0) \\
& ParaLS$\star$(ours)  & \textbf{18.5(16.8)} & \textbf{41.0(35.4)} & \textbf{52.1(48.3)} & \textbf{79.5(75.0)}  & \textbf{64.1(57.8)} \\
\hline
\end{tabular}
\caption{Evaluation results of substitute generation and substitute ranking on LS07 and CoInCo datasets. The scores in parentheses are only calculated by the substitutes from the substitute generation step. The baselines are Word2Vec \cite{melamud2015modeling}, BERT \cite{zhou2019bert}, BERT+WordNet \cite{michalopoulos2022lexsubcon}, GR-RoBERT \cite{lin-etal-2022-improving}, and XLNet+WordVec \cite{arefyev2020comparative,seneviratne-etal-2022-cilex}. Best values are bolded.}
\label{LS07andCoInCo}
\end{table*}

\textbf{Metrics.} For evaluating LS07 and CoInCo, we use the official metrics "best", "best-m", "oot", "oot-m" in SemEval 2007 task as well as Precision@1 (P@1) as our evaluation metrics, following the previous LS methods \cite{zhou2019bert,michalopoulos2022lexsubcon}. Among them, "best", "best-m" and "P@1" evaluate the quality of the best predictions, while both "oot" (out-of-ten) and "oot-m" evaluate the coverage of the gold substitute candidate list by the top 10 predictions.

In SwordS, a word is regarded as $acceptable$ if it is judged to be good by more than five out of ten annotators, and $conceivable$ if selected by at least one annotator. For the evaluation metrics, the authors \cite{lee-etal-2021-swords} use the harmonic mean of the precision and recall given the gold and top-10 system-generated substitutes. As gold substitutes, they use either the acceptable or conceivable words, and calculate the corresponding scores $F_a$ and $F_c$, respectively.

\textbf{Baselines.} We compare our methods ParaLS and ParaLS$\star$ with the following baselines, Word2Vec \cite{melamud2015simple}, BERT \cite{zhou2019bert}, BERT+WordNet \cite{michalopoulos2022lexsubcon}, GR-RoBERT \cite{lin-etal-2022-improving}, and XLNet+Word2Vec \cite{arefyev2020comparative,seneviratne-etal-2022-cilex}. Arefyev et al. \cite{arefyev2020comparative} linearly combine the prediction of pretrained language models XLNet and Word2Vec. Afterward, Seneviratne et al. \cite{seneviratne-etal-2022-cilex} adopt four features to rank the substitutes generated by XLNet and Word2Vec.

\textbf{Implementation Details.} To implement an English paraphraser, we fine-tune BART-base\footnote{https://dl.fbaipublicfiles.com/fairseq/models/bart.base.tar.gz} model in fairseq. The initial learning rate is set to $lr=3 \times 10^{-5}$ and dropout is set to 0.1. We adopt the Adam optimizer with $ \beta_{1}=0.9 $, $ \beta_{2}=0.999 $, $ \epsilon = 10^{-8}$. We choose an English paraphrase dataset ParaBank2 \cite{hu-etal-2019-large} to train the paraphraser. In our experiments, we duplicate all the samples by exchanging source sentence and target sentence. We use the BLEURT large model\footnote{https://huggingface.co/Elron/bleurt-large-512} for the calculation of BLEURT score.  BARTScore fine-tuned on ParaBank2 can be downloaded here\footnote{https://github.com/neulab/BARTScore}. We use the LS07 dev set for tuning the hyperparameters in our model. The weights for the prediction score (Paraphraser), BARTScore, and BLEURT for ParaLS and ParaLS$\star$ are 0.02, 1, and 1, respectively. The number of outputted paraphrases $K$ is set to 50. The lookahead length of ParaLS$\star$ is 2. Following the existing work \cite{zhou2019bert,michalopoulos2022lexsubcon,seneviratne-etal-2022-cilex}, only the top 10 substitutes are used for evaluation.

\subsection{Experimental Results}

\begin{table}[]
\centering
\begin{tabular}{l|cc}
\hline
\textbf{Method} &$F_a$&$F_c$\\\hline
GPT3&22.7&36.3\\
WORDTUNE&23.5&34.7\\\hline
BERT &17.2&27.5\\
XLNet+Word2Vec&21.7(19.9)&34.5(31.5)\\
\cdashline{1-3} 
ParaLS&23.5&38.6\\
ParaLS$^*$&\textbf{24.9(22.8)}&\textbf{40.1(37.0)}\\
\hline
         
\end{tabular}
\caption{Results on SwordS dataset. The results of two commercial systems GPT3 \cite{brown2020language} and WordTune \cite{AI21} are from Lee et al. \cite{lee-etal-2021-swords}. For all metrics, the higher, the better.}
\label{swords}
\end{table}

\textbf{Comparison of LS methods.} The results of our methods as well as the state-of-the-art methods on LS07 and CoInCo are displayed in Table \ref{LS07andCoInCo}. Typically, performance is evaluated by selecting the top substitutes after executing the substitute ranking step. To exclude the influence of substitute ranking, we also present the results without substitute ranking in parentheses.

As can be observed, our methods, ParaLS and ParaLS$\star$, demonstrate superior performance over the latest LS methods (GR-RoBERT and XLNet+Word2Vec) across all metrics in the LS07 and CoInCo datasets. Notably, ParaLS$\star$ without the step of substitute ranking outperforms all baselines, including the best baseline XLNet+Word2Vec, which utilizes four features for substitute ranking. ParaLS$\star$ without substitute ranking significantly outperforms ParaLS without substitute ranking, which means that the decoding with lookahead heuristic in ParaLS$\star$ is very useful.

The results on SwordS are presented in Table \ref{swords}. Our method ParaLS and ParaLS$\star$ achieve the highest $F_a$ score and $F_c$ score, largely outperforming the best baseline XLNet+Word2Vec as well as two commercial methods GPT-3 and WordTune. Unlike GPT-3, which is fine-tuned by a prompt-based learning framework from multiple samples of the development set in SwordS, ParaLS and ParaLS$\star$ do not rely on any LS dataset.

In comparison to LS methods based on pretrained language models, our methods possess the following three advantages: 

(1) The paraphraser has been specifically trained to learn lexical variations. This could give it an advantage over pre-trained language models, which are generally trained on a wide range of tasks and may not be as focused on lexical substitution.

(2) The paraphraser is better at preserving the original meaning and context of the text, as it has been specifically designed to rewrite text while maintaining its meaning. This could be particularly important for lexical substitution tasks, as the goal is often to find substitutions that are semantically similar to the target word.

(3) The paraphraser can generate more diverse or varied substitutions. Pre-trained language models, on the other hand, are more general-purpose and may not be as adept at generating diverse substitutions.

\begin{table}\centering
\begin{tabular}{l|cc}\hline
\textbf{Method} &   \textbf{LS07} & \textbf{CoInCo} \\\hline
 ParaLS$\star$(Ours)    &  \textbf{65.2}  & \textbf{60.0} \\
 -w/o BARTScore  & 63.6 & 59.1 \\
 -w/o BLEURT  & 64.1 & 58.9\\
 -w/o Paraphraser  & 64.5 & 59.2\\
 o. Paraphraser &61.9  & 57.5\\
 o. BARTScore &62.8  & 57.4 \\
 o. BLEURT &59.5   & 55.3 \\
ParaLS(Ours)    &  65.1  & 60.0 \\
 \cdashline{1-3} 
 XLNet+Word2Vec & 60.5 & 55.6 \\
 BERT+WordNet &  60.6 & 58.0 \\
 CILex3 & 57.8 & 53.6 \\
 BERT&  58.6 & 55.2 \\
 Word2Vec  & 55.1  & 50.2 \\ 
  \hline
\end{tabular}
\caption{Comparison of GAP scores (\%) in the substitute ranking sub-task. The results from XLNet+Word2Vec \cite{arefyev2020comparative}, CILex3 \cite{seneviratne-etal-2022-cilex} are presented. "-w/o" indicates a ParaLS framework without the specific feature. "o." indicates that only one specific ranking feature is used.  }
\label{ranking}\end{table}

\textbf{Comparison of Substitute Ranking.} We also evaluate our substitute ranking strategies on both the LS07 and CoInCo datasets. In this sub-task of LS task, assume that the substitute candidates are provided, each method aims to create the most appropriate ranking of the candidates. Following prior work \cite{zhou2019bert,michalopoulos2022lexsubcon}, we use GAP score \footnote{https://tinyurl.com/gap-measure} for evaluation in the subtask, which is a variant of MAP (Mean Average Precision). We also output the results of the proposed method ParaLS$\star$ by removing one feature or two features.

The results are displayed in Table \ref{ranking}. XLNet+Word2Vec, BERT+WordNet, and CILEX3 utilize 2, 4, and 4 features respectively to rank the substitutes, which include Gloss-sentence similarity score, sentence similarity score, and WordNet similarity score, among others. Our results obtained solely by using the BARTScore or Paraphraser feature surpass those of the baselines, with BARTScore exhibiting particularly strong performance. BLEURT also demonstrates superior performance when compared to CILEX3 and BERT. These results confirm that text generation evaluation metrics (BARTScore or BLEURT) are better suited for substitute ranking than prior methods. The performance of ParaLS$\star$ when one feature is removed demonstrates that all the features have a positive impact on the performance of ParaLS$\star$.

The proposed strategy using BARTScore or BLEURT for ranking substitutes based on the change of the sentence's meaning after embedding them into the original sentences is likely effective because it directly addresses the primary goal of lexical substitution, which is to preserve the meaning of the original sentence while replacing a word. By using text generation evaluation metrics such as BARTScore and BLEURT to compute the relationship between the original and updated sentences, the method can quantify the extent to which the meaning of the original sentence has been preserved by each substitute.

\begin{table}
\centering
\begin{tabular}{l|ccccc}\hline
 & best & b.m & oot & o.m & P@1 \\\hline
ParaLS$\star$  & \textbf{18.5} & \textbf{41.0} & \textbf{52.1} & \textbf{79.5} & \textbf{64.1}\\
-w/o Pa. & 17.8  & 39.7 &  51.4 & 79.0  & 61.2 \\
-w/o BA.  & 17.83 & 39.0  & 51.3 & 78.1  & 62.1 \\
-w/o BL. & 17.4  & 38.2 & 50.2  & 77.8 & 60.6\\
o. Pa. & 16.4 & 35.6 & 48.3 & 75.1 & 57.9 \\
o. BA. & 15.7 & 34.4 & 48.2 & 75.6  & 55.3 \\
o. BL.  & 14.9  & 31.3  & 48.2  & 74.3 & 53.4 \\\hline
\end{tabular}
\caption{Ablation study of ranking features for ParaLS$\star$ on CoInco dataset.  "-w/o" indicates ParaLS$\star$ without the specific feature. "o." indicates that only one specific ranking feature is used. "Pa.", "BA.", "BL.", "b.m" and "o.m" are denoted as "Paraphrase", "BARTScore", "BLEURT", "Best.m" and "oot.m", respectively.}
\label{ablation}
\end{table}

\textbf{Ablation Study.} To further evaluate the impact of each ranking feature on the performance of our method, we conducted an ablation study on ParaLS$\star$. The results are presented in Table \ref{ablation}. Both BARTScore and BLEURT are observed to be beneficial in enhancing the performance of ParaLS$\star$. The ablation study, by isolating and testing the performance of individual features, illustrates that the Paraphraser feature alone achieves the best performance, thereby highlighting the effectiveness of our decoding with lookahead heuristics in generating high-quality substitutes. 

\begin{table}
\centering
\begin{tabular}{l|l}
\hline
\textbf{Inst1} &  inauguration of \textbf{free} zone in $\cdot\cdot\cdot$\\ \hline
Labels  & open,unrestricted,unlimited, $\cdot\cdot\cdot$ \\
\hline
 BERT &safe,\textcolor{red}{open},public,reserve,reserved \\
\cdashline{2-2} 
 \multirow{2}{*}{XLNet} &complimentar,\textcolor{red}{open},exclusive, \\
 & new,digital  \\
 \cdashline{2-2} 
 \multirow{2}{*}{ParaLS} & \textcolor{red}{open},liberty,fair,\textcolor{red}{unrestricted}, \\
 & \textcolor{blue}{liberated} \\
 \cdashline{2-2} 
 \multirow{2}{*}{ParaLS$\star$ } &\textcolor{red}{open},liberty,\textcolor{blue}{autonomous},\\
 & \textcolor{blue}{independent},\textcolor{red}{unrestricted} \\
\hline
\textbf{Inst2} & i just hope they \textbf{keep} me here \\ \hline
Labels  & retain,stash,leave,hold,guard,$\cdot\cdot\cdot$ \\
\hline
 BERT & have,want,get,bring,take \\
 \cdashline{2-2} 
 XLNet &  \textcolor{blue}{maintain},\textcolor{blue}{stay},\textcolor{red}{hold},stick,have\\
 \cdashline{2-2} 
 ParaLS & \textcolor{red}{hold},\textcolor{red}{leave},\textcolor{blue}{stay},lock,have \\
 \cdashline{2-2} 
 ParaLS$\star$ &\textcolor{red}{hold},\textcolor{red}{leave},have,\textcolor{blue}{stay},\textcolor{blue}{put} \\
\hline
\textbf{Inst3} & $\cdot\cdot\cdot$ pulled out a  secret \textbf{code} for $\cdot\cdot\cdot$ \\ \hline
Labels  &  encryption,signal,password, $\cdot\cdot\cdot$\\
\hline
 \multirow{2}{*}{BERT} & \textcolor{blue}{combination},key,sequence, \\
 & message,number \\
 \cdashline{2-2} 
 \multirow{2}{*}{XLNet} &\textcolor{red}{password},message,key, \\
 & address,number  \\
 \cdashline{2-2} 
 \multirow{2}{*}{ParaLS} & \textcolor{red}{password},\textcolor{blue}{cipher},\textcolor{red}{encryption}, \\
 & message,protocol \\
 \cdashline{2-2} 
 \multirow{2}{*}{ParaLS$\star$} &\textcolor{red}{password},\textcolor{blue}{cipher},\textcolor{red}{encryption}, \\
 & protocol,message \\
\hline
\textbf{Inst4} & $\cdot\cdot\cdot$   drop an \textbf{atomic} bomb $\cdot\cdot\cdot$ \\ \hline
Labels  & nucleus,molecule,ion \\
\hline
 \multirow{2}{*}{BERT} & bomb,element,atmosphere, \\
 & \textcolor{red}{nucleus},uranium\\
  \cdashline{2-2} 
XLNet & earth,world,universe,planet,sun \\
 \cdashline{2-2} 
 \multirow{2}{*}{ParaLS} &\textcolor{blue}{nuclear},electron,\textcolor{red}{nucleus}, \\
 & particle,bomb \\
  \cdashline{2-2} 
 \multirow{2}{*}{ParaLS$\star$} &\textcolor{red}{nucleus},\textcolor{blue}{nuclear},bomb, \\
 & electron,electrons \\
\hline
\multirow{2}{*}{\textbf{Inst. 5}} &i do it as \textbf{somebody} who   \\ 
&  who has a conscience $\cdot\cdot\cdot$ \\
\hline
Labels&someone,one,person,anyone,$\cdot\cdot\cdot$\\
\hline
 \multirow{2}{*}{BERT} & \textcolor{red}{someone},\textcolor{blue}{anybody},\textcolor{red}{person},\\
 &\textcolor{red}{anyone},somewhere\\
 \cdashline{2-2} 
 \multirow{2}{*}{XLNet} &\textcolor{red}{someone},\textcolor{red}{person},persons, \\
 & somewhere,\textcolor{red}{one}\\
 \cdashline{2-2} 
 \multirow{2}{*}{ParaLS} & \textcolor{red}{someone},\textcolor{red}{one},\textcolor{red}{anyone},\textcolor{red}{person}, \\
 & \textcolor{blue}{anybody}\\
 \cdashline{2-2} 
 \multirow{2}{*}{ParaLS$\star$} &\textcolor{red}{someone},\textcolor{red}{one},\textcolor{red}{anyone},\textcolor{red}{person},\\
 & \textcolor{blue}{anybody}\\
\hline
\end{tabular}
\caption{The top five substitutes of five instances in CoInCo by LS methods. The target word is bolded, the substitutes in labels are marked in red, and the suitable substitutes not in labels are marked in blue.  }
\label{instances}
\end{table}

\textbf{Case Study.} To quantitatively evaluate the effectiveness of the substitutes generated by LS methods, we present five instances of CoInCo for analysis. Table \ref{instances} displays the top five generated substitutes. Upon examination, we find that many suitable substitutes, marked in blue, are not present in the Labels. As the labels are human-annotated, it is not possible to provide all suitable substitutes for each target word. We posit that the actual performance of ParaLS and ParaLS$\star$ is superior to the results computed by the metrics.

Furthermore, we see that our methods generate more high-quality substitutes than the baselines. Even when the methods generate unsuitable substitutes, the changes to the semantic information of the sentence are minimal. In the future, our methods could be utilized to enhance the coverage of substitutes in existing LS datasets.

\section{Conclusions}

We introduce two novel paraphraser-based LS methods named ParaLS and ParaLS$\star$, which generate substitute candidates by considering the context and preserving the sentence's meaning. Specifically, we design two decoding strategies that center on lexical variations of the target word during decoding and propose a substitute candidate ranking strategy by utilizing the newest text generation evaluation metrics. Experimental results show that ParaLS and ParaLS$\star$ significantly outperform the state-of-the-art LS methods. In the future, we will apply the methods to different languages, and verify our method on many downstream tasks to investigate further the method's general availability. 

\section*{Limitations}

Our method depends on a large-scale paraphrasing corpus. We only test our method on the English LS task.
Excluding English, other languages have large-scale paraphrasing datasets available, e.g., French, German, Chinese, Spanish, etc. Our method can be easily extended to these languages. But, for some languages that cannot obtain enough paraphrasing datasets, our proposed method cannot be used. Another limitation is that our method may struggle to generate substitutions for rare or unusual words and phrases, as they may not have encountered sufficient examples of these words in the training paraphrase data.

\section*{Ethics Statement}

One potential ethical consideration related to a LS method based on a paraphraser is the potential for biased or unfair language generation. If the training data used to develop the paraphraser is biased in some way (e.g., it disproportionately represents certain groups of people or uses certain words and phrases in a biased manner), this could lead to biased substitutions being generated by the model. It is important to ensure that the training data used to develop the model is diverse and free of bias in order to minimize the potential for unfair or biased language generation.

Another ethical consideration is the potential for the LS method to be used for malicious purposes, such as creating fake or misleading content. It is important to consider the potential consequences of the LS method's outputs and to put safeguards in place to prevent the LS method from being used for nefarious purposes.

\section*{Acknowledgement}
This research is partially supported by the National Natural Science Foundation of China under grants 62076217 and 61906060, and the Blue Project of Yangzhou University.

\bibliography{acl2023}
\bibliographystyle{acl_natbib}

\appendix 

\section{ Appendix A (More Experiments for Ablation Study)}

\textbf{1. The baselines.} We compare our methods ParaLS and ParaLS$\star$ with the following baselines. 

(1) Word2Vec: The words that have the highest similarities are selected as substitute candidates from the word embedding modeling whose vectors are closer in terms of cosine similarity with the target word \cite{melamud2015simple}. 

(2) BERT: BERT proposed by \cite{zhou2019bert} applies dropout to the embedding of the target word for partially obscuring the target word. 

(3) BERT+WordNet: Michalopulos et al. \cite{michalopoulos2022lexsubcon} integrated the knowledge from WordNet into the embedding of BERT. 

(4) GR-RoBERT: Lin et al. \cite{lin-etal-2022-improving} proposed an auxiliary gloss regularizer module to BERT pre-training, to enhance word semantic similarity.

(5) XLNet+Word2Vec\cite{arefyev2020comparative,seneviratne-etal-2022-cilex}:  \cite{arefyev2020comparative} linearly combines the prediction of pretrained language models XLNet and Word2Vec. Afterward, Seneviratne et al. \cite{seneviratne-etal-2022-cilex} adopt four features to rank the substitutes generated by XLNet and Word2Vec.

\textbf{2. Influence of different ranking features.} In the paper, we give the results of CoInCo datasets. Here, we give the results of LS07 datasets. The results are shown in Table \ref{ablation2}. The conclusions are consistent with the conclusions of CoInCo.

\begin{table}
\centering
\begin{tabular}{|l|ccccc|ccccc}\hline
 & best & b.m & oot & o.m & P@1 \\\hline
ParaLS$\star$  & \textbf{24.0} & \textbf{42.0} & \textbf{60.5} & \textbf{79.3} & \textbf{58.8} \\
-w/o Pa. & 22.2 &  38.9 & 58.5 & 76.8 & 54.4  \\
-w/o BA. & 23.6 & 40.9 & 59.6 & 78.0 & 57.3 \\
-w/o BL. & 23.7 & 41.4 & 59.1 & 78.5  & 58.0\\
o. Pa. & 22.3 & 39.0 & 57.3 & 76.1 & 54.3 \\
o. BA. & 20.2 & 35.0 & 55.8 & 75.3 & 50.5 \\
o. BL. & 18.6 & 30.0 & 54.9 & 70.7  & 46.7 \\\hline
\end{tabular}
\caption{Ablation study of ranking features for ParaLS$\star$  on LS07 dataset.  "-w/o" indicates ParaLS$\star$ without the specific feature. "o." indicates that only one specific ranking feature is used.}
\label{ablation2}
\end{table}

\begin{figure}[htbp] %插入图片
\subfloat[]{%
\pgfplotsset{compat=1.3}
\begin{tikzpicture}[scale=0.45,baseline] %tikz图片
\begin{axis}[
    xlabel=Lookahead Length,
    ylabel=$F_a$,
    label style={font=\Large},
    tick align=outside, %刻度在外显式
    legend style={at={(0.6,0.25)},anchor=north,font=\large} 
    ]

\addplot[sharp plot,mark=*,blue,densely dotted] plot coordinates { 
    (0,19.3)
    (1,22.6)
    (2,22.8)
    (3,22.9)
    (4,22.8)
    (5,22.9)
};
\addlegendentry{ParaLS$\star$(w/o ranking)}
\addplot[sharp plot,mark=triangle,red] plot coordinates {
    (0,23.5)
    (1,24.6)
    (2,24.5)
    (3,24.8)
    (4,24.9)
    (5,24.8)
};
\addlegendentry{ParaLS$\star$}
\end{axis}
\end{tikzpicture}}
\subfloat[]{%
\pgfplotsset{compat=1.3}
\begin{tikzpicture}[scale=0.45,baseline] %tikz图片
\centering
\begin{axis}[
    xlabel=Lookahead Length, %横坐标名
    ylabel=$F_c$, %纵坐标名
    label style={font=\Large},
    tick align=outside, %刻度在外显式
    legend style={at={(0.6,0.25)},anchor=north,{font=\large}} %图例在图下方显示
    ]
\addplot[sharp plot,mark=*,blue,densely dotted] plot coordinates { 
    (0,30.3)
    (1,36.5)
    (2,36.8)
    (3,37.1)
    (4,37.0)
    (5,37.1)
};
\addlegendentry{ParaLS$\star$(w/o ranking)}
\addplot[sharp plot,mark=triangle,red] plot coordinates {
    (0,38.6)
    (1,40.2)
    (2,40.0)
    (3,40.2)
    (4,40.1)
    (5,40.1)
};
\addlegendentry{ParaLS$\star$}
\end{axis}
\end{tikzpicture}}

\caption{Effect of varying lookahead length for ParaLS$\star$. (a) the results using metric $F_a$, and (b) the results using metric $F_c$.}
\label{fig:lookahed}
\end{figure}
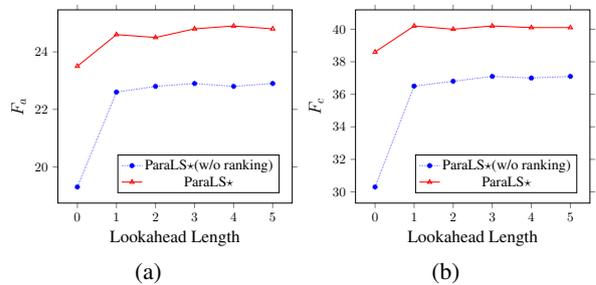

\begin{table}[h]
\centering
\begin{tabular}{l|l}
\hline
\textbf{Method} &\textbf{Runtime(s)}\\\hline
BERT &1.00\\
XLNet+Word2Vec&3.56\\
\cdashline{1-2} 
ParaLS&1.05\\
ParaLS$^*$ w/o ranking &  1.6 \\
ParaLS$^*$&1.96\\
\hline
\end{tabular}
\caption{The average running time of one instance. }
\label{runningtime}
\end{table}

\begin{table*}
\centering
\begin{tabular}{|l|l|ccccc|}\hline
\textbf{Dataset} & \textbf{Method} & best & best-m & oot & oot-m & P@1 \\\hline
\multirow{5}{*}{LS07} 
& XLNet+Word2Vec   & 23.3 & 40.9 & 56.3 & 74.8 & 55.9\\
& ParaLS (Transformer)  & \textbf{24.1} & \textbf{42.4} & 58.2 & 76.5 & 58.3\\
& ParaLS$\star$(Transformer)  & \textbf{24.1} & \textbf{42.2} & 59.4 & 77.4 & 58.6 \\
& ParaLS(BART) & 23.5 & 41.5 & 59.0 & 77.9 & 56.9 \\
& ParaLS$\star$(BART)  & 24.0 & 42.2 & \textbf{60.5} & \textbf{79.3} & \textbf{58.8}\\
\hline
\multirow{5}{*}{CoInCo} & 
XLNet+Word2Vec & 16.4 & 35.8& 46.9 & 73.0& 57.3\\
& ParaLS(Transformer) & 18.1 & 40.0 & 49.2 & 75.4 & 62.6 \\
& ParaLS$\star$(Transformer)  & 18.2 & 40.4 & 50.3 & 76.7 & 63.3 \\
& ParaLS(BART) & 18.1 & 40.1 & 50.7 & 78.1 & 62.4 \\
& ParaLS$\star$(BART)  & \textbf{18.46} & \textbf{40.96} & \textbf{52.14} & \textbf{79.48} & \textbf{64.11}\\
\hline
\end{tabular}
\caption{Results of ParaLS and ParaLS$\star$ using two different paraphrasers on LS07 and CoInCo datasets. For comparison, we also show the results of the best baseline "XLNet+Word2Vec" \cite{seneviratne-etal-2022-cilex}. }
\label{Abl_paraphraser}
\end{table*}

\textbf{3. Influence of lookahead length.} ParaLS$\star$ has a parameter of lookahead length. In this experiment, we will analyze the influence of lookahead length on the performance of ParaLS$\star$. We vary the length of the lookahead from 0 to 5. When lookahead length equals 0, ParaLS$\star$ is transformed into ParaLS.

The results are displayed in Figure \ref{fig:lookahed}.  We see that the performance ParaLS$\star$ is robust when varying lookahead length.

\textbf{4. The running time of LS method.} We give the average running time of one instance in Table \ref{runningtime}. We run 100 instances in CoInCo dataset, and compute the average time of one instance.

We see that ParaLS only need 1.05 second for one instance, close to BERT\cite{zhou2019bert}. XLNet+Word2Vec \cite{seneviratne-etal-2022-cilex} is the slowest LS method.

\textbf{5. Influence of different paraphrasers.} We do these experiments to verify the influence of different paraphrasers on the performance of ParaLS. In our paper, we adopt pretrained modeling BART to fine-tune an English paraphraser. Here, we train a Transformer model in FairSeq with a 6-layer encoder and decoder, 512-dimensional embeddings, 8 encoder-decoder attention heads, and 0.1 dropout. The initial learning rate is set to $lr=3 \times 10^{-4}$. We adopt the Adam optimizer with $ \beta_{1}=0.9 $, $ \beta_{2}=0.999 $, $ \epsilon = 10^{-8}$. 

The results are shown in Table \ref{Abl_paraphraser}. We see that the performance of our proposed ParaLS and ParaLS$\star$ is not significantly affected by the specific paraphrase model that is used.

\section{Appendix B (Case Study)}
\label{sec:appendix}

Here, we give the generated top 10 substitutes of 10 instances in CoInCo to analyze the generated substitutes by our method (ParaLS and ParaLS$\star$) and the baselines (BERT \cite{zhou2019bert} and XLNet \cite{seneviratne-etal-2022-cilex}.

\begin{table*}
\centering
\begin{tabular}{l|l}
\hline
\textbf{Inst. 1} & Chron editors note : each week , the chronicle offers readers a look at the more unusual fruits,\\ &\textbf{vegetables} and herbs of each season and how to use them .  \\ \hline
Labels  & veggies;produce;vegetable specimen;plant;herbage;\\
\hline
 BERT & foods;spices;grains;crops;berries;grasses;beans;\textcolor{red}{plants};shrubs;ingredients \\
 XLNet & herbs;crops;flowers;onions;foods;\textcolor{red}{plants};grains;seeds;fruits;potatoes\\
 ParaLS & \textcolor{red}{greens};\textcolor{red}{plants};foodstuffs;crops;\textcolor{red}{veggies};\textcolor{blue}{veg};seeds;varieties;vines;cereals\\
 ParaLS$\star$& \textcolor{blue}{greens};\textcolor{red}{plants};crops;\textcolor{red}{veggies};varieties;foodstuffs;seeds;\textcolor{blue}{veg};\textcolor{red}{produce};vines  \\
\hline\hline

\textbf{Inst. 2} &  If they continued to \textbf{resist} , he pulled out a secret code for their bosses . \\ \hline
Labels  & refuse;rebel;thwart the matter;stonewall;refrain;oppose;object;disbelieve;defy;decline;\\
&counteract;be uncooperative;abstain; \\
\hline
 BERT &  \textcolor{red}{refuse};\textcolor{blue}{struggle};\textcolor{blue}{protest};\textcolor{red}{rebel};submit;comply;obey;reject;flee;escape\\
 XLNet & \textcolor{red}{refuse};\textcolor{blue}{protest};persist;comply;react;hesitate;evade;obey;respond;submit\\
 ParaLS & \textcolor{red}{refuse};\textcolor{red}{oppose};\textcolor{blue}{protest};\textcolor{blue}{struggle};fight;evade;\textcolor{red}{rebel};object;deny;revolt\\
 ParaLS$\star$&\textcolor{red}{refuse};\textcolor{red}{oppose};fight;\textcolor{blue}{struggle};\textcolor{blue}{protest};evade;\textcolor{red}{rebel};object;deny;\textcolor{red}{defy} \\
\hline\hline

\textbf{Inst. 3} &He 's a right handed bat , which complements Palmeiro off the \textbf{bench} .\\\hline
Labels&wood;wait area;stand;seat;reserve;replacement;relief;pine;dugout;box;bleacher;backup;\\
&auxiliary\\
\hline
 BERT & field;pitch;team;plate;bat;opener;start;rest;ball;squad \\
 XLNet & lineup;mound;team;roster;plate;field;diamond;\textcolor{red}{box};spot;bullpen\\
 ParaLS & \textcolor{red}{stand};field;court;bleachers;\textcolor{red}{dugout};plate;pitch;mound;ground;line\\
 ParaLS$\star$& field;\textcolor{red}{stand};court;pitch;deck;\textcolor{red}{box};\textcolor{red}{dugout};plate;mound;\textcolor{red}{bleachers}\\
\hline\hline

\textbf{Inst. 4} &Grande dame of cooking still going strong at 90 : Julia Child \textbf{celebrates} in san francisco\\ \hline
Labels&rejoice;party;enjoy;dance\\
\hline
 BERT & celebrations;celebration;sings;remembers;promotes;holidays;performs;starts;wins;promotions \\
 XLNet & celebration;celebrations;holidays;festivities;holiday;feast;birthday;shows;parade;festival\\
 ParaLS &\textcolor{blue}{commemorates};is;presents;gala;festivities;\textcolor{blue}{commemorate};anniversary;\textcolor{red}{party};birthday;glorifies\\
 ParaLS$\star$& \textcolor{blue}{commemorates};\textcolor{red}{rejoice};feast;\textcolor{red}{rejoices};feasts;\textcolor{blue}{cheers};festivities;\textcolor{red}{dances};revels;presents\\
\hline\hline

\textbf{Inst. 5} &Responsible seafood \textbf{sales} are the catch of the day\\ \hline
Labels&purchase;transaction;vending;purchasing;deal;buying;barter\\
\hline
 BERT & \textcolor{red}{purchases};selling;markets;prices;vendors;trading;buyers;stores;products;donations \\
 XLNet & \textcolor{red}{purchases};selling;sellers;markets;marketing;shipments;prices;buyers;businesses;retailers\\
 ParaLS & \textcolor{red}{purchases};\textcolor{blue}{sells};selling;exports;products;sold;prices;markets;sell;\textcolor{red}{deals}\\
 ParaLS$\star$&\textcolor{red}{purchases};\textcolor{blue}{sells};deliveries;exports;\textcolor{red}{transactions};products;selling;supplies;markets;\textcolor{red}{deals}\\
\hline\hline

\end{tabular}
\caption{The top 10 substitutes of five instances in CoInCo using LS methods. The target word is bolded, the substitutes in labels are marked in red, and the suitable substitutes not in labels are marked in blue. Here, the baselines are BERT \cite{zhou2019bert} and XLNet \cite{seneviratne-etal-2022-cilex}. "XLNet+Word2Vec" is abbreviated as XLNet.  }
\label{sample1}
\end{table*}

\begin{table*}
\centering
\begin{tabular}{l|l}
\hline

\textbf{Inst. 6} &Sony corp. completed its tender offer for Columbia pictures entertainment inc., with Columbia \\
&shareholders \textbf{tendering} 99.3\% of all common shares outstanding by the Tuesday deadline.\\ \hline
Labels&pay;offer;issue;give;get;earn;deal\\
\hline
 BERT & \textcolor{blue}{bid};auction;submit;deposit;present;surrender;broker;ballot;forward;dispatch \\
 XLNet &\textcolor{red}{offering};\textcolor{blue}{bidding};submitting;selling;taking;placing;securing;buying;\textcolor{blue}{providing};accepting\\
 ParaLS &\textcolor{red}{offering};submitting;\textcolor{blue}{bidding};accepting;bid;requesting;receiving;\textcolor{red}{giving};\textcolor{blue}{providing};buying\\
 ParaLS$\star$&\textcolor{red}{offering};\textcolor{blue}{bidding};\textcolor{blue}{submitting};bid;accepting;\textcolor{red}{giving};\textcolor{blue}{proposing};\textcolor{blue}{providing};requesting;holding\\
\hline\hline

\textbf{Inst. 7}& " We 've discontinued \textbf{selling} swordfish , chilean seabass , orange roughy and marlin , "\\\hline
Labels&offering;vend;serve;peddling;distributing\\\hline
 BERT & \textcolor{blue}{marketing};\textcolor{red}{offering};buying;producing;sales;retail;sale;\textcolor{blue}{trading};shipping;export\\
 XLNet &buying;sales;marketing;sale;\textcolor{red}{offering};purchasing;\textcolor{blue}{trading};promoting;shipping;\textcolor{blue}{supplying}\\
 ParaLS & \textcolor{blue}{retailing};\textcolor{blue}{marketing};\textcolor{blue}{trading};\textcolor{red}{distributing};\textcolor{red}{offering};sellin;sales;trafficking;sale;\textcolor{blue}{serving}\\
 ParaLS$\star$&\textcolor{blue}{marketing};\textcolor{blue}{trading};\textcolor{blue}{retailing};\textcolor{red}{distributing};buying;\textcolor{red}{offering};\textcolor{blue}{supplying};delivering;\textcolor{red}{peddling};\\
 &\textcolor{blue}{merchanting}\\
\hline\hline

\textbf{Inst. 8} &The federal complaint offers many \textbf{details} of the alleged conspiracy , including excerpts from \\
&a transcript of the Italian wiretaps .\\
\hline
 Labels&specific;point;fact;tidbit;snippet;item;issue;facet;count;account\\\hline
 BERT &  outlines;\textcolor{blue}{information};descriptions;\textcolor{blue}{specifications};highlights;documents;features;stories;\textcolor{red}{facts};\\
 &terms\\
 XLNet &descriptions;\textcolor{blue}{information};outlines;elements;aspects;\textcolor{red}{facts};highlights;\textcolor{red}{accounts};components;\\
 &features\\
 ParaLS &\textcolor{blue}{particulars};\textcolor{blue}{aspects};\textcolor{red}{specifics};\textcolor{red}{facts};\textcolor{blue}{information};\textcolor{blue}{evidence};elements;\textcolor{blue}{indications};\textcolor{blue}{clarifications};\\
 &\textcolor{red}{facets}\\
ParaLS$\star$&\textcolor{blue}{particulars};\textcolor{blue}{aspects};\textcolor{red}{specifics};descriptions;\textcolor{red}{facts};elements;\textcolor{blue}{evidence};\textcolor{blue}{indications};\textcolor{blue}{information};\\
&\textcolor{red}{facets}\\
\hline

\textbf{Inst. 9} &The \textbf{new} factory , which will begin normal production early next year , will employ about 1,000 \\
&people .\\\hline
Labels&late;most recent;recent;projected;pristine;future;fresh;expect;come;added\\\hline
 BERT &  rebuilt;expanded;\textcolor{blue}{upcoming};planned;expanding;proposed;combined;large;larger;second\\
 XLNet & \textcolor{red}{future};modern;\textcolor{red}{latest};proposed;first;planned;expanded;current;original;main\\
 &features\\
 ParaLS & \textcolor{red}{fresh};young;fellow;rookie;\textcolor{blue}{incoming};\textcolor{red}{recent};\textcolor{blue}{next};\textcolor{blue}{emerging};younger;own\\
ParaLS$\star$&\textcolor{blue}{next};\textcolor{red}{fresh};\textcolor{red}{latest};\textcolor{red}{future};\textcolor{blue}{emerging};\textcolor{red}{recent};novel;innovative;production;construction\\
\hline

\textbf{Inst. 10} &Electronic theft by foreign and \textbf{industrial} spies and disgruntled employees is costing \\
&U. S. companies billions and eroding their international competitive advantage .\\\hline
Labels  & business;trade;mechanized;manufacturing;industrialized;economic \\
\hline
 BERT &  industry;\textcolor{red}{manufacturing};corporate;\textcolor{blue}{commercial};technical;multinational;\textcolor{blue}{technological};factory;\\
 &internal;chemical\\
 XLNet & industry;corporate;domestic;\textcolor{blue}{commercial};internal;institutional;international;regional;national\\
 &independent;\\
 ParaLS & \textcolor{blue}{commercial};\textcolor{red}{manufacturing};corporate;factory;\textcolor{red}{business};\textcolor{red}{economic};sectoral;professional;\\
 &\textcolor{blue}{technological};international \\
 ParaLS$\star$& \textcolor{red}{manufacturing};\textcolor{blue}{commercial};\textcolor{red}{business};corporate;factory;professional;\textcolor{red}{economic};sectoral;\\
 &\textcolor{blue}{technological};international\\
\hline\hline

\end{tabular}
\caption{The top 10 generated substitutes of five instances in CoInCo using LS methods. The target word is bolded, the substitutes in labels are marked in red, and the suitable substitutes not in labels are marked in blue. Here, the baselines are BERT \cite{zhou2019bert} and XLNet \cite{seneviratne-etal-2022-cilex}. "XLNet+Word2Vec" is abbreviated as XLNet.  }
\label{sample1}
\end{table*}

\end{document}